\documentclass[letterpaper, 10 pt, conference]{ieeeconf}  %

\IEEEoverridecommandlockouts                              %

\makeatletter
\let\NAT@parse\undefined
\makeatother

\makeatletter
\let\labelindent\undefined
\makeatother

\usepackage{subcaption}
\usepackage{graphics} %
\usepackage{epsfig} %
\usepackage{mathptmx} %
\usepackage{times} %
\usepackage{amsmath} %
\usepackage{amssymb}  %
\usepackage{booktabs} %
\usepackage{multirow} %
\usepackage{color}
\usepackage{enumitem} %
\usepackage[sort,square,numbers]{natbib}
\usepackage{url}
\usepackage{xspace}   %
\usepackage{placeins}         %

\definecolor{julieta_colour}{RGB}{117,112,179} %
\definecolor{shenlong_colour}{RGB}{27,128,80} %
\definecolor{raquel_colour}{RGB}{217,95,2}     %
\definecolor{andrei_colour}{RGB}{231,101,101}
\definecolor{sasha_colour}{RGB}{102,166,30}    %

\DeclareMathAlphabet{\mathcal}{OMS}{cmsy}{m}{n}

\definecolor{citecolor}{RGB}{34,139,34}
\usepackage[breaklinks=true,letterpaper=true,colorlinks, citecolor=citecolor,bookmarks=false]{hyperref}

\newcommand{\eg}{e.g.\@\xspace}	%
\newcommand{\etal}{\emph{et al.\@\xspace}}
\newcommand{\lidar}{LiDAR}
\DeclareMathOperator*{\argmin}{\arg\!\min}

\newcommand{\ba}{\mathbf{a}}

\newcommand{\bp}{\mathbf{p}}

\newcommand{\bn}{\mathbf{n}}

\newcommand{\bz}{\mathbf{z}}

\newcommand{\cZ}{{\mathcal{Z}}}

\newcommand{\cD}{\mathcal{D}}

\newcommand{\cG}{\mathcal{G}}

\newcommand{\by}{\mathbf{y}}

\title{\LARGE \bf
Pit30M: A Benchmark for Global\\ Localization in the Age of Self-Driving Cars
}

\author{
Julieta Martinez$^{1}$, Sasha Doubov$^{1, 2}$, Jack Fan$^{1}$, \\
Ioan Andrei B\^{a}rsan$^{1,3}$, Shenlong Wang$^{1,3}$, Gell\'{e}rt M\'{a}ttyus$^{1}$, Raquel Urtasun$^{1,3}$ \\
\thanks{$^{1}$Uber Advanced Technologies Group} %
\thanks{$^{2}$University of Waterloo}
\thanks{$^{3}$University of Toronto}
}

\begin{document}

\maketitle
\thispagestyle{empty}
\pagestyle{empty}

\begin{abstract}
We are interested in understanding whether retrieval-based localization
approaches are good enough in the context of self-driving vehicles.
Towards this goal, we introduce Pit30M, a new image and \lidar{} dataset with over 30 million frames,
which is 10 to 100 times larger than those used in previous work.
Pit30M is captured under diverse conditions (i.e., season, weather, time of the day, traffic), and provides accurate localization ground truth.
We also automatically annotate our dataset with historical weather and astronomical data, as well as
with  image and \lidar{} semantic segmentation as a proxy measure for occlusion.
We  benchmark multiple existing methods for image and \lidar{} retrieval and,
in the process, introduce a simple,  yet effective convolutional network-based \lidar{} retrieval method that is competitive with the state of the art.
Our work provides, for the first time, a benchmark for sub-metre retrieval-based localization at city scale.

The dataset, its Python SDK, as well as more information about
the sensors, calibration, and metadata, are available on the project website:
\begin{center}
  \url{https://pit30m.github.io/}
\end{center}

\end{abstract}

\section{Introduction}

\begin{figure*}[t]
    \includegraphics[width=1\linewidth,trim=0mm 0mm 0mm 0mm,clip=true]{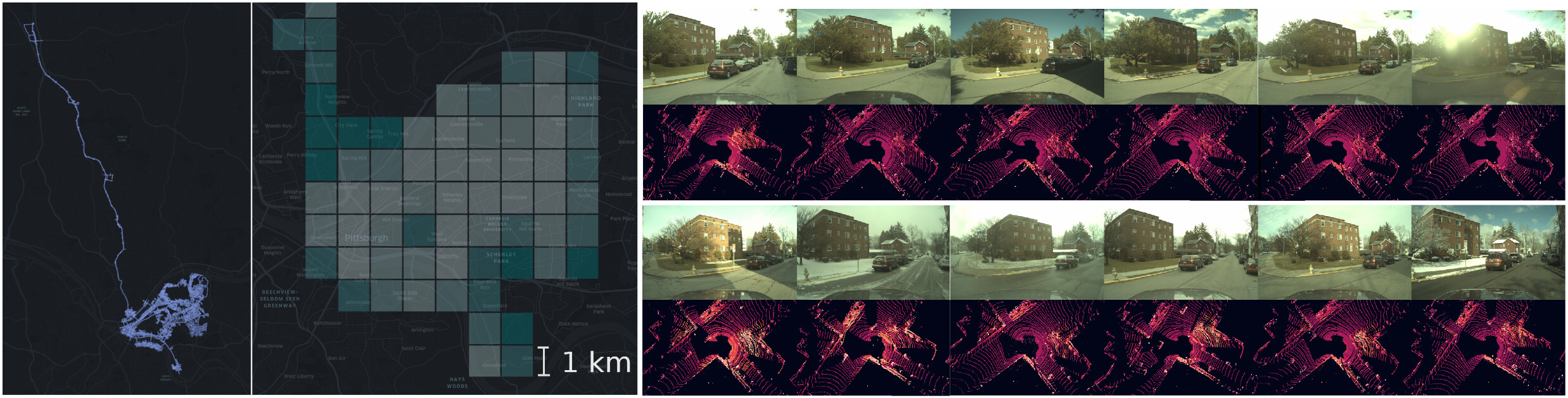} %
    \vspace{-1.5em}
    \caption{
        \textbf{Our new localization dataset, Pit30M.}
        Left: Each square is 1 km$^2$, for a total area of about 50 km$^2$ plus over 20 km of highway in the Pittsburgh Metropolitan Area.
        Right: Examples of image and \lidar{} point clouds taken in the same place at different times.
    }
    \label{fig:extent}
\end{figure*}

Localizing an autonomous agent accurately and in real time is a fundamental
problem in robotics. In the context of autonomous driving, it allows
the self-driving vehicles (SDVs), to navigate to their destinations.
Furthermore, accurate localization enables the use of HD maps, which boosts and contextualizes downstream autonomy tasks such as perception, motion forecasting, and motion planning.

Localization tasks can be divided into two broad categories: online
localization and global localization.
Online localization assumes that the pose at the previous time step is known,
and is tasked with propagating that information over time, as well as combining it
with current sensory measurements.
However, small errors may accumulate during online localization, making the
pose estimate drift over time or even fail altogether.
Global localization aims to overcome this issue,
  as it re-estimates the global pose without assumptions on previous steps.
Hence, global localization is an important fail-safe module that allows self-driving vehicles to recover %
from temporary online localization failures.

Autonomous driving faces unique challenges when it comes to global localization.
To systematically evaluate various global localization approaches
in this context, we need a benchmark that reflects the
setting and its particular challenges.
Ideally, the dataset employed to carry out this study should be diverse,
large-scale, and have accurate ground truth over a variety of
environments and traffic scenarios.
Furthermore, since autonomous driving platforms typically carry a
variety of sensors that provide complementary information (such as LiDAR, camera,
GPS, and IMU), the benchmark should contain multi-sensory data to enable
researchers to exploit multi-modal inputs for the global localization task.
Unfortunately, no existing dataset fulfills all these criteria.

In this paper we introduce Pit30M, a dataset that spans over a year of driving in
Pittsburgh, PA, USA,
comprising over 1\,000 trips, 25\,000 km, and 1\,500 hours driven under different times of day, seasons, and diverse weather conditions.
Moreover, we provide accurate ground truth poses (under 10 cm of error) for all our data.
With over 30 million images and \lidar{} sweeps,
our dataset is one to two
orders of magnitude larger than the biggest publicly-available dataset for this
task.
We also provide metadata such as time of day, weather, and approximate occlusion by
leveraging image and \lidar{} segmentation,
which allow us to formally quantify the diversity in our dataset and understand localization errors.
We give an overview of the geographical and temporal extent of our dataset in Figure~\ref{fig:extent}.
The dataset, available at \href{https://pit30m.github.io}{pit30m.github.io} can be accessed using our
\href{https://github.com/pit30m/pit30m}{open-source Python SDK}. The SDK is available under the
MIT license, while the dataset is licensed under the Creative Commons Attribution-NonCommercial-ShareAlike 4.0 International License.

We investigate both image- and \lidar{}-based approaches to retrieval localization.
For visual localization, and with our dataset's scale, diversity and density, we find that a modern convolutional backbone with a simple pooling scheme perform on par with state-of-the-art architectures specifically designed for this task, such as NetVLAD~\cite{netvlad}.
For \lidar{}-based localization, we investigate both the latest network architectures and suitable pointcloud representations.
We show that bird's-eye view voxelization coupled with a strong convolutional backbone is competitive with the best previously proposed
pointcloud representations and architectures for this task -- which also rely on NetVLAD pooling.
Finally, we provide an analysis of the failure modes and complementarity of \lidar{}- and camera-based localization. %

\section{Related work}

Structure- and retrieval-based localization have been studied in the computer vision community for decades.
We summarize these two areas of research, as well as regression-based and hybrid approaches to localization.
We also briefly review datasets typically used for this task.

\subsection{Structure-based localization}

\noindent {\bf Image structure-based localization:}
Cameras perceive a 2d projection of the 3d scene, so research in this area
has focused on building consistent 3d maps of the world~\cite{agarwal2009building}.
Given a query image it is possible to recover its pose by
matching its 2d features against the 3d model, and
then solving the perspective-n-point (PnP) problem.
Towards this goal, researchers have explored a variety of 2d-to-3d descriptor matching
techniques~\cite{dusmanu2019d2, irschara2009structure, sattler2011fast, li2012worldwide, sattler2015hyperpoints, liu2017efficient}.

While these approaches can be very accurate, several drawbacks remain.
Scalability (maintaining a very large 3d database), for example, remains challenging;
while fine vocabularies~\cite{sattler2015hyperpoints, havlena2014vocmatch} and
model compression~\cite{camposeco2019hybrid, Lynen2019} provide
ways to accelerate matching in large scenes, accuracy suffers, and
building and storing 3d models at city scale requires large engineering efforts.
Building systems that are robust to long-term changes~\cite{sattler2018benchmarking} also remains an active area of research.

\vspace{0.2cm}
\noindent {\bf \lidar{} structure-based localization:}
\lidar{}s provide a 3d point cloud that can be aggregated over time to provide a dense 3d reference scan of
a scene.  This scan can later be used for localization by aligning new observations using, \eg,
local registration methods such as iterative closest point~\cite{yoneda2014lidar,pomerleau2015review},
or 2d template matching~\cite{Levinson2007,Levinson2010,RyanW.2014,deepgill,Wei2019}.

A substantial portion of work on \lidar{}-based localization focuses on online localization.
Levinson~\etal~\cite{Levinson2007} proposed one of the first \lidar{}-based online localization systems capable of
centimetre-accurate localization, and subsequent work has improved the robustness of such systems by using
probabilistic maps~\cite{Levinson2010,RyanW.2014}, leveraging deep learning to bypass the need for calibrated
intensity~\cite{deepgill}, or incorporating real-time kinematic (RTK) information~\cite{Wan2017}.
Nevertheless, given their online nature, such systems require highly accurate initialization
(assumed to be provided by another system), and rely on dense high-definition maps
which
can be prohibitively expensive to collect and build at scale.

\subsection{Retrieval-based localization}
Retrieval approaches to localization do not rely on a pre-built map,
but assume access to a database with localized sensor observations.
The pose of a query can thus be estimated by finding the nearest observation in the database.
Since database entries may be represented by a single vector, these approaches tend to be more scalable;
however, their accuracy is limited by the density and coverage of the underlying database, and
finding compact yet discriminative representations remains difficult.

\vspace{0.2cm}
\noindent {\bf Image retrieval-based localization:}
Classical methods extract local invariant features~\cite{sift, surf}, %
and aggregate them into a global descriptor such as %
visual bag-of-words~\cite{filliat2007visual} or VLAD~\cite{vlad, densevlad}.
Candidate re-ranking and geometric verification are sometimes used as a second
stage to further boost performance~\cite{knopp2010avoiding, sattler2017large}.
Recent work has used deep convolutional neural networks (CNNs) to learn compact
visual representations~\cite{netvlad, radenovic2016cnn, rmac}.
For instance, NetVLAD~\cite{netvlad} uses a CNN and differentiable VLAD pooling
to learn global image representations for retrieval in an end-to-end manner,
and RMAC \cite{rmac}~builds a compact deep feature vector with Region of Interest (RoI) pooling.

\vspace{0.2cm}
\noindent {\bf \lidar{} retrieval-based localization:}
Compared to images, \lidar{}-based retrieval methods remain relatively unexplored.
While handcrafted 3d descriptors have been used for 3d registration and recognition tasks~\cite{tombari2010unique, rusu2009fast},
we are not aware of classical global pooling techniques applied to \lidar{} retrieval-based localization.

Recently,
work has concentrated on learning deep descriptors from 3d point clouds~\cite{he2016m2dp, dewan2018learning, Klokov_2017_ICCV}.
PointNetVLAD~\cite{pointnetvlad} uses
PointNet~\cite{qi2016pointnet} to generate local per-point features,
which are then aggregated by a VLAD~\cite{netvlad} layer.
PCAN~\cite{zhang2019pcan} improves upon PointNetVLAD by learning an attention map for aggregation,
using an architecture inspired by PointNet++~\cite{qi2017pointnet++}.
Finally, LPD-Net~\cite{lpdnet} achieves state-of-the-art retrieval results using a graph neural network
to leverage local structure when learning global descriptors.
While these methods yield excellent results, they operate directly on raw point clouds, which is computationally expensive in general.

\subsection{Other localization approaches}

\noindent {\bf Regression-based localization:}
In these approaches, the model directly outputs a position in a known scene
\cite{shotton2013scene, posenet, brachmann2016uncertainty, brachmann2018learning}.
The most significant advantage of these methods is that they allow for
localization without access to external databases, leading to low memory usage
and fast inference.
While promising, these methods have been shown to not generalize well to
city-scale localization~\cite{sattler2018benchmarking, sattler2019understanding}.

\vspace{0.2cm}
\noindent {\bf Hybrid Localization:}
Other methods take components from both retrieval- and map-based localization,
or incorporate non-traditional elements in their map representation and inference.
For instance, recent work has explored the complementarity of semantics, temporal information, and
global pose regression~\cite{valada2018deep, vlocnet++, schonberger2018semantic},
or local and global localization features~\cite{sarlin2019coarse} in a single model.
Another recent line of work has explored ways of learning map representations that
are better suited for visual localization tasks~\cite{henriques2018mapnet,
brahmbhatt2018geometry}.%

\subsection{Current localization datasets}
Datasets are a key component of research in large-scale localization.
On one hand, datasets that span large city areas such as SFO-Landmarks~\cite{chen2011city}
and Tokyo/Pittsburgh Street view~\cite{densevlad, torii2013visual}
often provide only GPS readings as reference poses~(SFO-Landmarks also considers visual overlap to compute ground truth).
Unfortunately, GPS can be inaccurate by several metres,
making it hard to quantify the error of localization methods that aim for sub-metre accuracy.
This is evident in the evaluation protocol of most previous work in large-scale retrieval-based localization,
where a database match is considered correct if it is within 20 or 25 metres of the
query~\cite{densevlad, netvlad, pointnetvlad, zhang2019pcan, lpdnet}.

Other datasets such as Cambridge~\cite{posenet} and Aachen~\cite{sattler2018benchmarking} derive ground truth from SfM models,
for which the error is hard to quantify and,
due to the computational cost of SfM, remain hard to extend to city-scale.
Urban driving datasets such as KITTI~\cite{geiger2012we}, Oxford RobotCar~\cite{maddern20171},
DeepLoc~\cite{vlocnet++} and NCLT~\cite{nclt} use robotic platforms for data collection.
However, they either only cover multiple areas of the environment just once,
or focus on revisiting the same route up to 100 times, limiting geographic
extent.
Finally, these datasets often derive ground truth localization from GPS and inertial filters,
which do not achieve centimetre-level accuracy. %

Recent work has used manual annotations to provide more accurate ground truth localization,
either via manual verification of 2d-3d matches with existing SfM models~\cite{sattler2017large},
or by manually aligning \lidar{} and SfM point clouds in publicly-available datasets~\cite{sattler2018benchmarking}.
However, these annotations have remained relatively small-scale efforts,
providing around 100\,000 localized images in the largest case.
In contrast, we aim for a dataset that provides millions of accurately-localized
images and \lidar{} point clouds.

\section{Pit30M: Global localization at city scale}
\label{sec:dataset}

\begin{table*}[!bt]
\centering
\footnotesize
\begin{tabular}{l rrrrrrrr}

    \toprule
    & Distance (km) & $^\dagger$Images & $^\dagger$Accurate GT & Geo span (km$^2$) & Time span & Sessions & (type) \lidar{} \\

    \midrule

    Pittsburgh 250k~\cite{torii2015visualPAMI}  & --        & 250       & --         & $\sim$ 16 & -- & -- & -- \\

    Tokyo 24/7~\cite{densevlad}                 & --        & $^*$600   & $^{**}$1   &  2.56  & -- &  -- & -- \\

    SFO Landmarks~\cite{chen2011city}           & --        & 1\,700    &  1\,700    & $\sim$ 18  & -- & -- & (Unspecified) \checkmark\\

    \midrule
    DeepLoc~\cite{vlocnet++}                    & 4         & 2         & 2     & 0.015    & 1 day & 10 & (Unspecified) \checkmark\\

    NCLT~\cite{nclt}                            & 147       & 630       & 630   & $\sim$ 1 & 15 months & 27 & (Velodyne 32) \checkmark\\

    \midrule

    Aachen~\cite{sattler2012image}              & --        & 5         & \cite{sattler2018benchmarking} 5  & $\sim$ 1.5 & -- & -- & -- \\

    CMU~\cite{bansal2014Localization}           & 98.7       & 82        & \cite{sattler2018benchmarking} 82 & -- & 3 months & 12 & --  \\

    Oxford robotcar~\cite{maddern20171}         & 1\,000    & $^{\ddagger}$8\,500 & \cite{sattler2018benchmarking} 38 & $\sim$ 10  & 18 months & 133 & (2x SICK LMS) \checkmark\\
    \midrule

    Pit30M (Ours)                               & 25\,000   & 30\,000 & 30\,000 & $\sim$ 50  & 14 months & 1343 & (Velodyne 64) \checkmark\\
    \bottomrule
\end{tabular}

\caption{\footnotesize {\bf Comparison of datasets for large-scale visual localization.}
$^\dagger$In thousands.
$^{\ddagger}$The dataset has over 20M images in total, but we consider only the frontal camera to make it comparable to our dataset.
$^*$Including synthesized views.
$^{**}$The number of query images localized manually.
}
\vspace{-1em}
\label{tab:datasets}
\end{table*}

\begin{figure}[!bt]
    \centering
    \includegraphics[width=\linewidth,trim=2mm 0mm 5mm 65mm,clip=true]{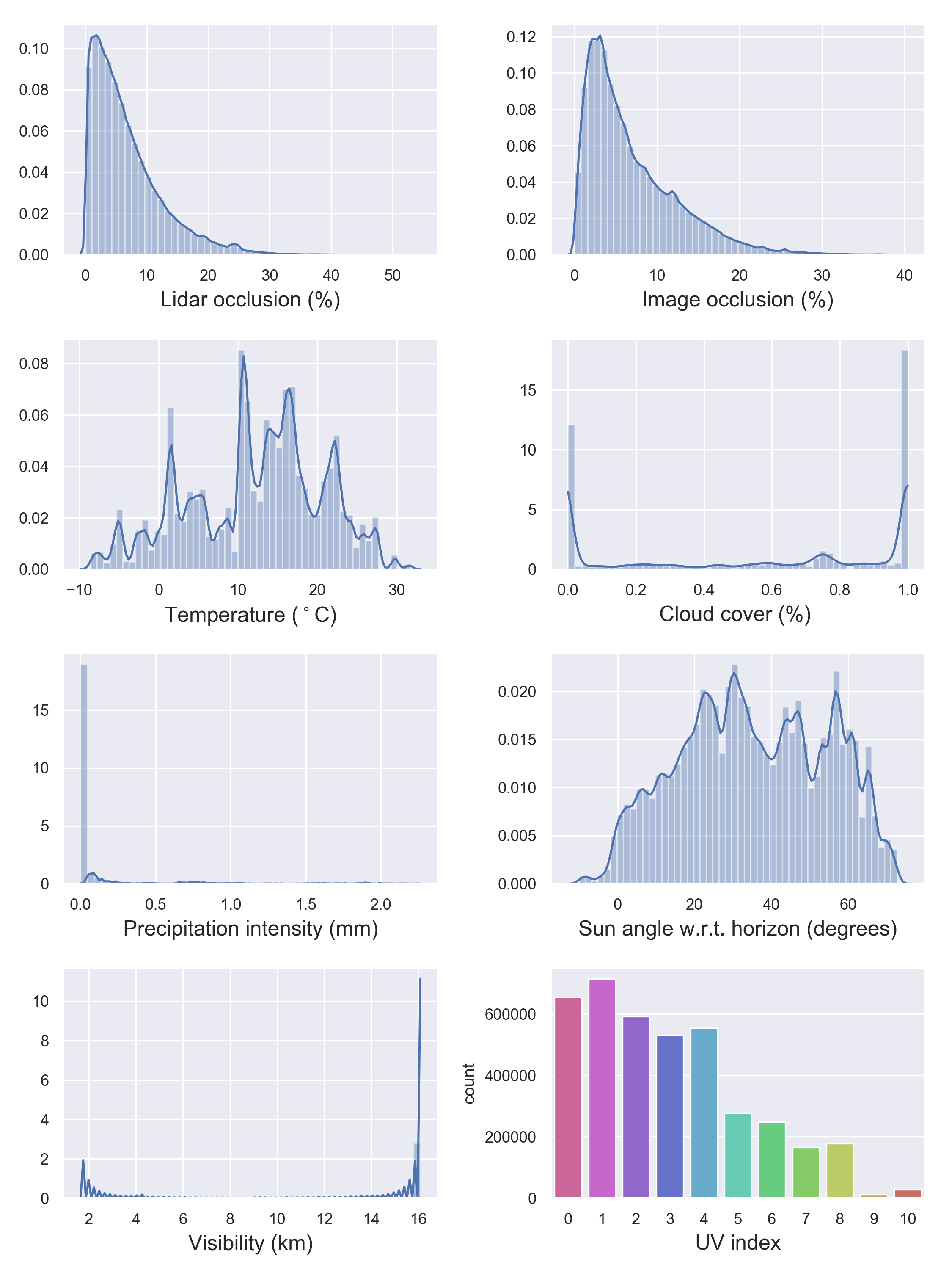}
    \caption{
        \textbf{Probability density functions (PDFs) for metadata in Pit30M.}
        For a complete description of these tags, please refer to the Appendix.
      }
    \label{fig:pdfs}
    \vspace{-1.5em}
\end{figure}

We assume that the region where our SDV is located has previously been covered with an appearance database.
This dataset should ideally have three characteristics:
\begin{enumerate}[wide, labelwidth=!, labelindent=0ex, leftmargin=1em, topsep=0.5em, itemsep=-0.5ex, partopsep=1ex, parsep=0.75ex]
\item {\bf Diversity} in appearance is necessary to train models that learn to recognize the same site under changes due to weather, seasons, illumination, construction, occlusion, and dynamic objects in the scene.
\item {\bf Scale} refers to the area spanned by the dataset. We want our
  dataset to cover an entire city, as this is the typical operational
  domain of a self-driving car.%
\item {\bf Accurate ground truth} provides a clear evaluation benchmark for methods that achieve sub-metre accuracy.
We can also use this ground truth as a supervisory signal to improve retrieval-based localization.
\end{enumerate}

We used our self-driving fleet to collect a dataset of 1\,343 trips and 30 million images and \lidar{} point clouds.
Our data was collected from Jan 2017 to Feb 2018 in the Metropolitan area of Pittsburgh, PA, USA.
Our vehicles carry a Velodyne HDL-64 \lidar{} sensor, a wheel odometer and an 
IMU, which we use to localize offline using vehicle dynamics and \lidar{} registration against a pre-existing dense 3d scan of the scene geometry.
These measurements are all fed to a commercial batch optimization system that has been validated to yield under 10cm  localization error.
We use an HD, global-shutter, colour camera
located in the roof of the vehicle, facing forward at all times, which provides images at a resolution of 1\,920 $\times$ 1\,200 pixels.
The horizontal and vertical fields of view are 78.6$^\circ$ and 52.5$^\circ$, respectively.
The intrinsic and extrinsic calibration parameters of the cameras and
\lidar{} (\eg{} the \lidar{}-to-camera rigid transformation) are computed
and validated a priori using a standard setup consisting in fiducial targets
and non-linear optimization.
We also carry a consumer-grade GPS sensor.
The continuous stream of
points produced by the \lidar{} is broken up into $100$ms partitions and
motion-compensated. 
The corresponding camera image is
selected such that it is as close as possible to the moment that the \lidar{}'s
rolling shutter passed through the middle of its FoV
The synchronization is within a few milliseconds.

Pit30M is, to the best of our knowledge, the largest benchmark for large-scale localization to date
both in terms of images, \lidar{} readings, and accurate ground truth
information.
Table~\ref{tab:datasets} provides summary statistics of existing datasets
(described in the previous section) as well as ours,
and Figure~\ref{fig:extent} shows the extent of our data. %
The proposed dataset
includes over 25\,000 km and 1\,500 hours of driving,
resulting in a benchmark that is one to two orders of magnitude larger than those used in previous work.
Moreover, our dataset spans all seasons, diverse weather conditions (including
rain, sleet, and snow), multiple times of day,
including  images taken at night and with low natural lighting, as well as construction and changes in buildings and pavement.

\vspace{0.2cm}
\noindent {\bf Large-scale metadata:}
Previous datasets have provided manual, trip-level metadata,
typically with the goal of identifying challenging conditions for localization.
For example, the Oxford dataset~\cite{maddern20171} provides 11 different tags including
``sun'', ``clouds'', ``dusk'', and ``snow'', and the CMU seasons
dataset~\cite{sattler2018benchmarking} includes tags for ``park'', ``urban'',
``foliage', and ``low sun'', among others.

Unfortunately,  trip-level tags can be ambiguous; \eg, the same trip may be sunny and cloudy at different times.
Instead, we have collected more granular metadata using historical weather and astronomical data, which can be obtained at scale.
In particular, we have collected weather via the \url{darksky.net} public API,
and estimated the angle of the sun in the sky using the \texttt{skyfield} library.
We have also used state-of-the-art \lidar{}~\cite{zhang2018efficient} and image~\cite{bai2017deep} semantic segmentation 
to estimate the degree of background occlusion in our dataset.
We showcase our labels by analyzing the results of our preliminary benchmark in Section~\ref{sec:experiments}.
Figure~\ref{fig:pdfs} shows probability density functions of some of our tags.

\section{Benchmarking large-scale localization}
\label{sec:study}

We turn our attention to benchmarking retrieval-based localization approaches.
Formally, let $\cZ$ be either an image or \lidar{} sweep point cloud,
let $\cG$ be the GPS pose
and $\mathbf{y}$ the position of the SDV we are trying to infer.

In the retrieval localization setting, we start from a dataset of pre-localized
sensor observations, and represent each full sensor reading as a vector $\bz = f(\cZ)$,
to obtain a database of vector-pose pairs
$\cD = \{(\bz_1, \by_1), (\bz_2, \by_2), \dots, (\bz_n, \by_n) \}$.
Given an online sensor reading $\cZ_q$,
we first compute the global feature representation $\bz_q = f(\cZ_q)$ and infer the current pose via (high-dimensional) nearest neighbour search:
\begin{equation}
	\hat{\bz} = \argmin_{\bz_i} \lVert \bz_q - \bz_i \rVert_2^2,
\end{equation}
and output the pose associated with the nearest dataset descriptor $\hat{\bz}$.
Since each observation is associated with a single vector $\bz_i$, the obtained pose is expressed in a global coordinate frame.

We now introduce a couple of
simple, yet effective retrieval convolutional networks for large-scale localization.
By leveraging the supervision provided by our dataset
(which is typical in an industrial setting),
we show that strong convolutional backbones
with simple pooling schemes can match the state of the art in image and \lidar{} retrieval.
This allows us to showcase the importance of and gains that are possible with our data,
and gives us insights into the state of the art of retrieval-based localization in the context of self-driving.

\begin{figure}[!tb]
	\centering
    \includegraphics[width=1\linewidth,trim=5mm 38mm 45mm 45mm,clip=true]{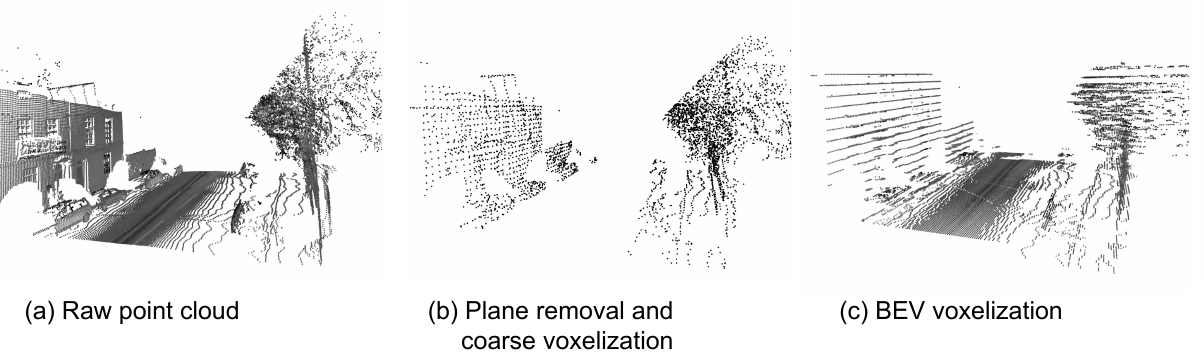}
	\caption{\textbf{\lidar{} representations benchmarked in this work.}
    (a) Raw point cloud (not used by any method).
    (b) Point cloud after ground plane removal and downsampling to 4\,096 points~\cite{pointnetvlad,zhang2019pcan,lpdnet}.
    (c) BEV voxelization with intensities.
	We use the latter as input to CNNs.}
	\label{fig:lidar}
	\vspace{-0.5em}
\end{figure}

\vspace{0.2cm}
\noindent {\bf Learning for retrieval:}
We rely on deep convolutional networks
trained with a standard triplet loss, common in the retrieval literature:
\begin{equation}
	\mathcal{L}_{\text{retrieve}} = \max \left \{ d(\ba, \bp) - d(\ba, \bn) + m, 0 \right \},
\end{equation}
where $d(\cdot, \cdot)$ is the Euclidean distance function, a triplet $(\ba, \bp, \bn)$ consists of three latent,
$\ell_2$-normalized embeddings produced by a network $f(\cZ)$.
In this context, $\ba$ is an ``anchor'' descriptor, $\bp$ is a ``positive'' descriptor and $\bn$ is a ``negative'' descriptor.
We build our triplets such that the geo-location of the positive image is
closer to the anchor than the negative image by a pre-defined margin
of at least $m = 0.5$ in embedding feature space.
In our experiment, we consider sensor readings within 1 metre to be positives, and within 2 and 4 metres to be negatives;
notice that this fine-grained distinction is enabled by the accurate
ground truth provided by our dataset.

We also make sure that the heading angle of the three images is within a range of $30^\circ$, so the images have overlapping fields of view.
Finally, we ensure that the positive and negative samples do not come from the
same trip as the anchor, which encourages the learned representations
to be invariant to factors such as time of day, weather, and dynamic objects in the scene.
We collect a triplet for each image in the dataset, and learn $f(\cdot)$ (in practice implemented as a Resnet-50~\cite{he2016deep}) via backpropagation.

\vspace{0.2cm}
\noindent {\bf GPS + retrieval:}
We also consider using  GPS  to restrict the search area.
This is a more realistic scenario in the context of self-driving, yet is less studied in the literature.
Here, we collect a set of embeddings located within $\tau$ metres from the GPS reading, where
$\tau$ is a tunable hyperparameter that we set based on the empirical error of
our GPS measurements ($\tau=20$m in our main results).
We then perform retrieval as in the global case, only restricted to this region.

\vspace{0.2cm}
\noindent {\bf Bird's-Eye View (BEV) representation:}
Although it is straightforward to use CNNs with images in the above formulation,
it is not immediately clear how that can be achieved when the input is a point cloud.
For this, we introduce a representation that is conceptually simple yet
achieves state-of-the-art results on publicly available benchmarks. %
(Please refer to the Appendix
for detailed experiments on the Oxford Robotcar~\cite{maddern20171} dataset.)
We preprocess the raw \lidar{} point cloud as a BEV multi-channel representation
by discretizing 3d space into an evenly-spaced voxelization of size $l \times w \times h$.
Crucially, we treat the resulting voxelization as a 2d image by discretizing the z axis into $c$ channels,
which can be plugged directly into standard 2d CNNs.
This representation has proven useful for efficient real-time \lidar{}
object detectors~\cite{chen2017multi, luo2018fast},
but here we show that it also produces competitive results on \lidar{} retrieval.
In contrast, previous retrieval work has operated directly on the point clouds,
which are heavily downsampled for computational reasons~\cite{pointnetvlad, lpdnet, zhang2019pcan}.
We visualize different \lidar{} representations in Figure~\ref{fig:lidar}.

\vspace{0.3cm}
\section{Experiments}
\label{sec:experiments}

\begin{table}[!bt]
    \centering
    \footnotesize

\begingroup
\setlength{\tabcolsep}{3pt} %

\newcommand{\best}[1]{\textbf{#1}}
\begin{tabular}{@{}l|rrrr||rr@{}}
\toprule
\% within (metres) & 0.25m & 0.50m & 1.0m & 5.0m & average & median \\
\midrule
GPS & 0.4 & 1.7 & 6.2 & 73.7 & 4.20 & 3.40 \\
\midrule
\midrule
{\bf Image-based methods} \\
VLAD                  & 8.59  & 20.01 & 33.44 & 51.40 & 2401.33 & 3.95\\
DenseVLAD             & 14.50 & 34.12 & 56.15 & 77.82 & 843.98 & 0.81\\
NetVLAD-OOB           & 13.85 & 32.47 & 55.27 & 82.38 & 476.33 & 0.84\\
NetVLAD               & 42.57 & 74.38 & 86.07 & 87.60 & 577.69 & 0.29\\
Resnet50 (Img)        & 45.40 & 78.51 & 90.39 & 91.87 & 418.64 & \best{0.27}\\
\midrule
GPS + VLAD            & 10.44 & 24.78 & 43.58 & 78.80 & 3.53 & 1.26\\
GPS + DenseVLAD       & 15.33 & 36.47 & 61.33 & 90.78 & 2.05 & 0.72\\
GPS + NetVLAD-OOB     & 14.47 & 33.88 & 58.24 & 90.87 & 2.05 & 0.77\\
GPS + NetVLAD         & 43.53 & 77.78 & 92.73 & 96.80 & 0.92 & 0.28\\
GPS + Resnet50 (Img)  & 45.74 & 79.79 & 93.49 & 97.06 & 0.86 & \best{0.27}\\
\midrule
\midrule
{\bf \lidar{}-based methods} \\
PointNet Max         & 36.53 & 66.46 & 78.82 & 80.77 & 944.28 & 0.34 \\
PointNetVLAD         & 51.60 & 83.29 & 91.89 & 93.16 & 322.85 & 0.24 \\
PCAN                 & 48.95 & 81.88 & 91.51 & 92.47 & 339.96 & 0.26 \\
BEV + Resnet50       & 60.17 & 86.08 & 91.39 & 92.56 & 353.27 & \best{0.20}\\
\midrule
GPS + PointNet Max   & 37.73 & 70.03 & 86.70 & 91.95 & 1.78 & 0.32 \\
GPS + PointnetVLAD   & 50.77 & 82.43 & 91.27 & 94.35 & 1.42 & 0.25 \\
GPS + PCAN           & 48.43 & 80.77 & 90.63 & 93.39 & 1.55 & 0.26 \\
GPS + BEV + Resnet50 & 59.38 & 84.88 & 91.25 & 93.74 & 1.51 & \best{0.21} \\
\bottomrule
\end{tabular}
\endgroup

    \caption{
    {\bf Detailed localization results for retrieval-based approaches.}
    We report the percent of correct predictions within different distance thresholds,
    and mean and median over the entire query set.
    Top: Image-based methods. Bottom: \lidar{}-based methods.}
    \label{tab:main-retrieval}
    \vspace{-1em}
\end{table}

\begin{figure}[!bt]
    \centering
    \includegraphics[width=1\linewidth,trim=0mm 0mm 0mm 0mm,clip=true]{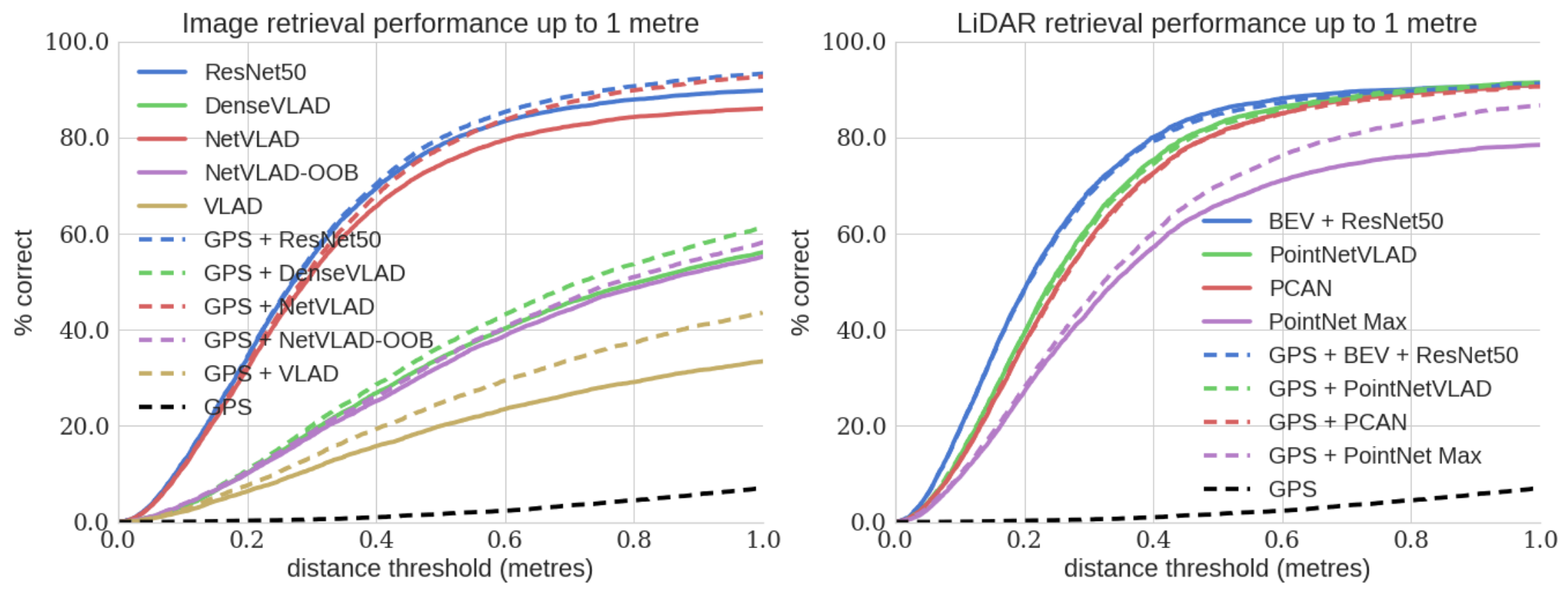}
    \caption{\textbf{Performance of retrieval-based methods.}
    Left: Image retrieval results.
    Right: \lidar{} retrieval results.\label{fig:main-retrieval}}
    \vspace{-0.2cm}
\end{figure}

\begin{figure*}[!bt]
    \centering
    \includegraphics[width=0.9\linewidth,trim=65mm 24.2mm 62mm 2mm,clip=true]{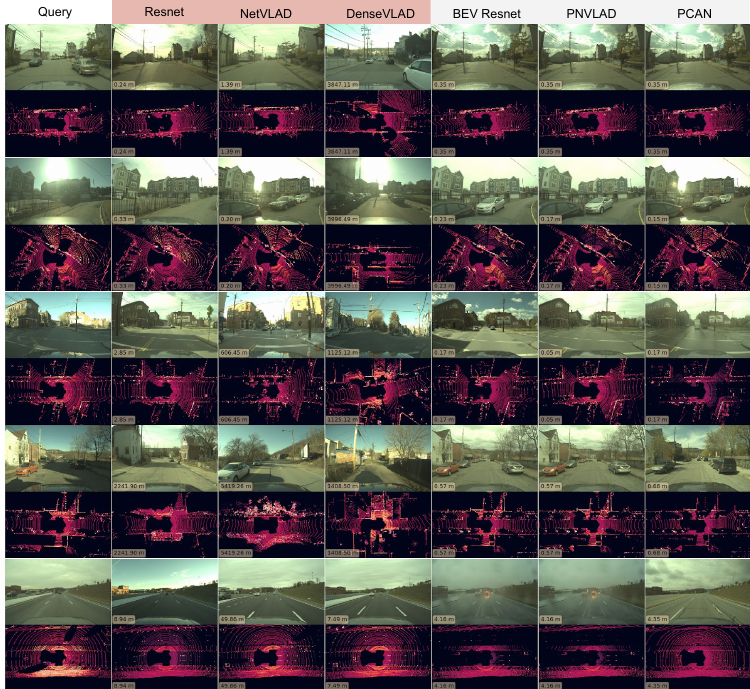}
    \caption{\textbf{Qualitative results under exhaustive search.}
        Left: Query. Middle: Image retrieval method. Right: \lidar{} retrieval methods.
    }
    \label{fig:a}
    \vspace{-0.3cm}
\end{figure*}

\subsection{Evaluation protocol}
We split Pit30M in terms of different trips.
This partition protocol suits the self-driving scenario well, where a fleet typically maps the drivable areas beforehand,
but new trips are faced with changes in visual appearance and new dynamic objects on the road.
We randomly select 941 trips for training, 134 for validation and 268 for testing.
We further select 10\,000 random query sensor readings from
the test partition
to report our final localization metrics.
We report the percentage of correctly localized queries for increasing distance ranges.
This results in a monotonically increasing curve for increasing distance,
with a hypothetical
perfect localizer having a performance of 100 for every distance value.

\subsection{Benchmarked methods}

\noindent {\bf Camera-based methods:}
From classical methods, we consider
SIFT-based VLAD~\cite{vlad} and DenseVLAD~\cite{densevlad}, a variant of VLAD specifically
designed for very dense datasets with high visual variability.
For these two methods,
we learn the visual vocabulary on a subset of 5M images from the training set, and use 128 SIFT clusters.
We also consider NetVLAD~\cite{netvlad}, a deep learning approach that uses
VGG-16 convolutional feature maps as local features,
and adds a learnable VLAD-based pooling layer.
We use the TensorFlow implementation of Cieslewski~\etal\footnote{\url{https://github.com/uzh-rpg/netvlad\_tf\_open}}~\cite{cieslewski2018data}.
Since previous work~\cite{sarlin2019coarse} has relied on the best model from~\cite{netvlad}
trained on Pittsburgh\footnote{\url{https://www.di.ens.fr/willow/research/netvlad/}} for global localization,
we also benchmark this pre-trained network out-of-the-box on our dataset,
so as to provide context about the performance of a strong baseline in the field.
We call this method NetVLAD-OOB.

\vspace{0.2cm}
\noindent {\bf \lidar{}-based baselines:}
We consider PointNet-Max \cite{qi2016pointnet},
PointNetVLAD~\cite{pointnetvlad}, and PCAN~\cite{zhang2019pcan}
and train them on the Pit30M dataset.
We use the publicly-available implementations of
PointNetVLAD\footnote{\url{https://github.com/mikacuy/pointnetvlad}}~\cite{pointnetvlad}.
We also implemented our own version of PCAN~\cite{zhang2019pcan}.
Note that we do not evaluate the state-of-the-art LPD-Net~\cite{lpdnet} on Pit30M due to the lack of a public implementation.
However, as shown on the Appendix,
benchmarking on the Oxford datasets shows that our BEV + Resnet50 method is
competitive with this strong baseline.

\subsection{Results}

\noindent {\bf Quantitative results:}
We show the results of our retrieval-based benchmark in Figure~\ref{fig:main-retrieval}, and detailed results in Table~\ref{tab:main-retrieval}.
Regarding image retrieval, while VLAD is outperformed by both DenseVLAD and NetVLAD--OOB,
the differences between the hand-crafted DenseVLAD and deep NetVLAD are rather small, and no clear winner emerges.
Our image-based Resnet50 baseline performs on par with NetVLAD.
In \lidar{} retrieval, BEV emerges as the best method overall, outperforming all its image counterparts and other \lidar{} methods.

\vspace{0.2cm}
\noindent {\bf Qualitative results:}
We show some qualitative results for the retrieval-based methods on Pit30m in Figure~\ref{fig:a}. %
We focus on challenging queries with glare, low sun angle, snow and rain.
For example, the second row shows a query with glare, where both ResNet and NetVLAD manage to localize correctly, but DenseVLAD fails.
The bottom row shows an interesting example where snow makes \lidar{} highly reflective; however all \lidar{} methods are able to localize
within $5$m.

\subsection{Analysis}

\begin{figure*}[!bt]
    \centering
    \includegraphics[width=0.9\linewidth,trim=0mm 0mm 0mm 0mm,clip=true]{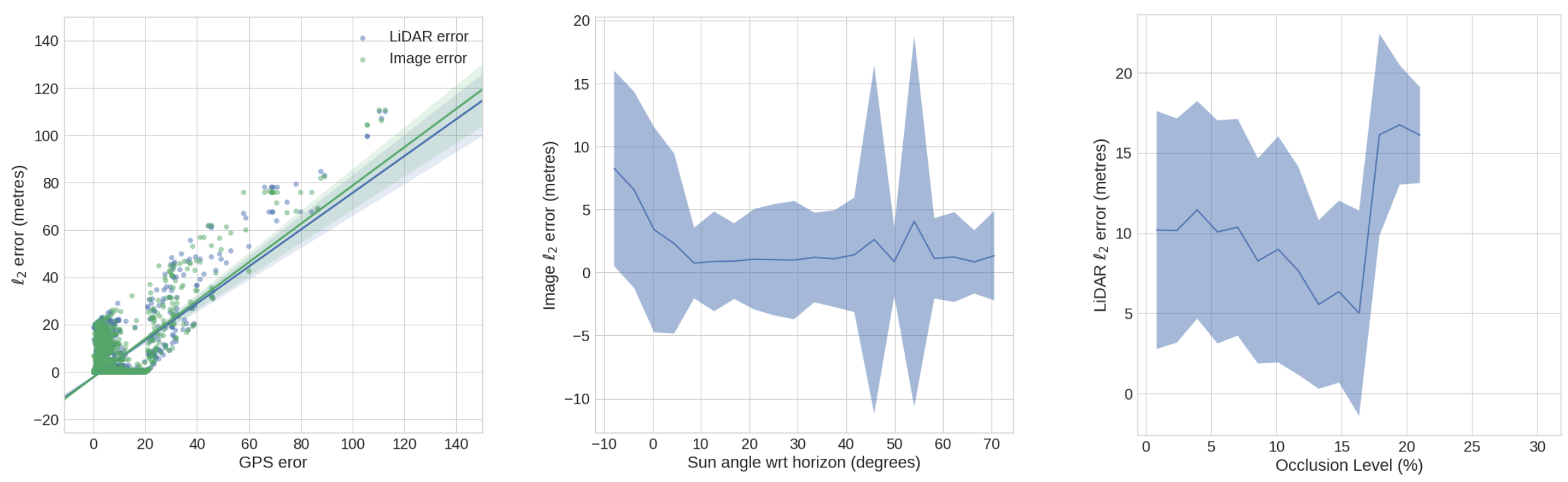}
    \caption{\textbf{Examples of analysis enabled by the Pit30M metadata.}
        Left: GPS error is correlated with both image and \lidar{} localization error.
        Middle: Image localization error vs.\ sun angle in the horizon
        (altitude angle). We
        observe a smooth error increase as the sun gets closer to the horizon.
        Right: We plot \lidar{} queries with more than 1 metre of error (failure cases) against \lidar{} occlusion.
        We observe a sharp spike in error when between 15 and 20\% of points are assigned to dynamic objects.
    }
    \label{fig:analysis}
\end{figure*}

\begin{figure}[!bt]
    \centering
    \includegraphics[width=1\linewidth,trim=0mm 0mm 0mm 0mm,clip=true]{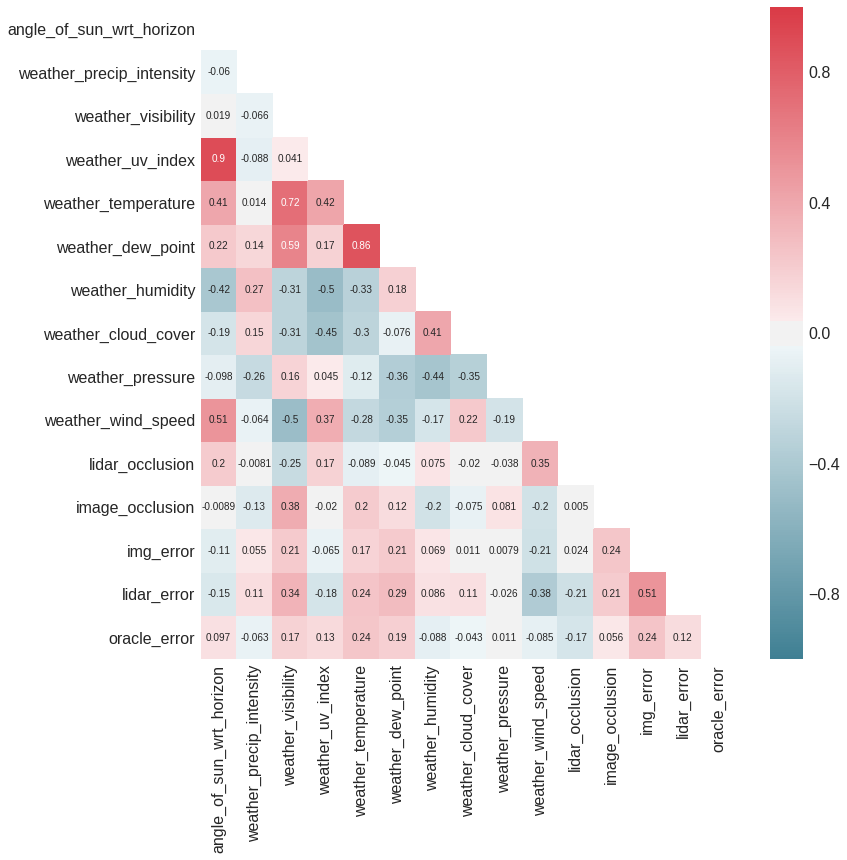}
    \caption{\textbf{Pairwise correlations between metadata in Pit30M and error of different methods.}
    "Oracle error" stands for a hypothetical method that is able to pick the best of either image or \lidar{} prediction for each query.
    }
    \label{fig:corr}
\end{figure}

We use the annotations provided by Pit30M to analyse the results of our  benchmarking.
We focus on understanding our best-performing methods,
GPS+Resnet-50 (images) and GPS+BEV+Resnet-50 (\lidar{}).

First, we observe in Figure~\ref{fig:analysis} (left), that large GPS error tends to cause
large overall localization errors for both images and \lidar{}. This is not surprising, since
we limit the search radius to $20$m around the GPS prediction.
In Figure~\ref{fig:corr}, we show the pairwise Pearson correlation coefficient between our labels and failure cases
of image and \lidar{} retrieval (a query is considered a ``failure'' if its
error exceeds $2$m),
after removing cases where GPS error is above $20$m.
While we observe that, for example, image error is correlated with image occlusion,
and image and \lidar{} errors are highly correlated,
this analysis is somewhat limited, as it is only able to capture linear correlations.
Thus, we further examine image error vs.\ sun angle in Fig.~\ref{fig:analysis} (middle);
here, we observe increased errors during dawn and twilight, and no effect during daytime.
We also examine \lidar{} error vs.\ occlusion, and observe a spike in errors when 15-20\% of points are assigned to
dynamic objects.

\section{Conclusions}

We have introduced Pit30M, a novel large-scale dataset for image and \lidar{} localization,
and studied retrieval-based methods in the context of self-driving cars.
Our dataset provides extensive metadata and sub-metre ground truth, and allows researchers to study accurate global localization at city-scale.
We have also provided an initial benchmark with multiple methods for visual and \lidar{} localization,
and in the process shown that strong modern convolutional backbones perform remarkably well in this scenario.
Our analysis also hints at future research directions using multi-sensor fusion, and highlights challenging scenarios
for localization.
Our dataset and metadata are available on the Pit30M project website. %

\section*{APPENDIX}
\addcontentsline{toc}{section}{Appendix}
\setcounter{section}{0}
\renewcommand{\thesection}{\Alph{section}}

\newcommand{\wrt}{w.r.t.}	%

\label{sec:experiments}

\section{Oxford Robotcar Experiments}

We benchmark different \lidar{} representations on the Oxford RobotCar dataset~\cite{maddern20171},
which contains over 100 trips and 1\,000 km driven in the city of Oxford, UK over the span of a year.
The vehicle carries two 2d \lidar{} sensors that scan the scene as the car moves through it.
This creates a 3D point cloud as shown in Fig~\ref{fig:lidar}.
Unfortunately, the dataset does not provide accurate ground truth for its sensor readings, as it only provides GPS locations.
Nonetheless, to the best of our knowledge, all previous work on \lidar{} retrieval-based localization has
benchmarked on this dataset~\cite{pointnetvlad, lpdnet, zhang2019pcan}.

We use the publicly available code by Uy~\etal~\footnote{\url{https://github.com/mikacuy/pointnetvlad}}~\cite{pointnetvlad}
to generate the training and test partitions on the dataset.
For each session, we aggregate the 2d lidar scans using vehicle dynamics to build a full reference map,
and associate them with global coordinates from the GPS readings.
Next, we generate train and test submaps by using two geographically disjoint sets of reference maps,
with each submap containing all the \lidar{} points collected over 20m of
driving.
For the training set, positive submaps are defined as being at most 10m from the query submap,
while negative submaps are defined as being at least 50m away from the query.

\noindent {\bf Inhouse datasets:}
Previous work has also typically benchmarked on the three ``inhouse'' datasets provided by Uy~\etal~\cite{pointnetvlad}.
Unfortunately, only heavily downsampled pointclouds are available online. %
We contacted the authors asking for the full pointclouds, but were told that they are no longer available,
so we cannot benchmark our method on them.

\noindent {\bf Evaluation protocol:}
We follow the standard evaluation procedure for Oxford RobotCar,
where a submap match must be within 25m of the query to be considered correct.
As such, the dataset contains 21\,711 submaps that are used for training and 3\,030 submaps for evaluation.
We generate the train and test submaps by splitting the corresponding reference maps at intervals of 10m and 20m, respectively.
For the baseline networks~\cite{qi2016pointnet, pointnetvlad, zhang2019pcan},
the raw point clouds are pre-processed by removing the ground planes from each submap,
and downsampled to 4\,096 points using a coarse voxel filter.
We follow previous work and report average recall@1, and average recall@1\%,
meaning whether a positive match is found within the top or top 1\%
of the retrieved maps respectively, averaged over all queries.

\noindent {\bf Training details:}
Our network is trained with a lazy quadruplet loss~\cite{pointnetvlad},
and each batch consists of one anchor, two positives and 18 negative examples.
Finally, we refresh the representation cache used for hard negatives every 1\,000 iterations.
We train our network using the Adam optimizer~\cite{kingma2014adam},
with an initial learning rate of 0.001, which we
decrease by 10$\times$ every time the validation accuracy plateaus.
Our network takes roughly 8 hours to train on a single 1080Ti GPU.

\begin{table}[!bt]
    \centering
    \small
    \begin{tabular}{l|rrr}
    \toprule
    & r@1\% & r@1 & Inference\\
    \midrule
    PointNet-Max~\cite{qi2016pointnet} & 73.87 & 54.16 & -- \\
    PointNetVLAD~\cite{pointnetvlad}   & 81.01 & 62.76 & 13.09 ms\\  %
    P-CAN~\cite{zhang2019pcan}         & 86.40 & 70.72 & -- \\
    LPD-Net light~\cite{lpdnetlight}   & 89.55 & 77.92 & 18.88 ms\\
    LPD-Net~\cite{lpdnet}              & \underline{94.92} & {\bf 86.28} & 23.58 ms\\
    \midrule
    BEV + Resnet50 (ours)              & {\bf 95.10} & \underline{86.13} & 8.41 ms\\
    \bottomrule
\end{tabular}

    \caption{
        {\bf Comparison of \lidar{}-based retrieval methods on the Oxford
        RobotCar dataset}.
        Our method achieves competitive results at lower inference times.
        All times benchmarked on an Nvidia 1080Ti GPU.
        \label{tab:robotcar}
    }
\end{table}

\noindent {\bf Results:}
We report results on Table~\ref{tab:robotcar}.
Our method is competitive with the state-of-the-art LPD-Net~\cite{lpdnet}, while being much faster at inference time.
Notably, our method achieves this by using a strong backbone, but without relying on NetVLAD pooling (unlike all our baselines).
We believe that this result shows the effectiveness of BEV + CNN representations for the task of \lidar{} retrieval-based
localization.

\section{Semantic labels in Pit30M}

\begin{figure*}[h]
    \centering
    \includegraphics[width=0.7\linewidth,trim=0mm 0mm 0mm 0mm,clip=true]{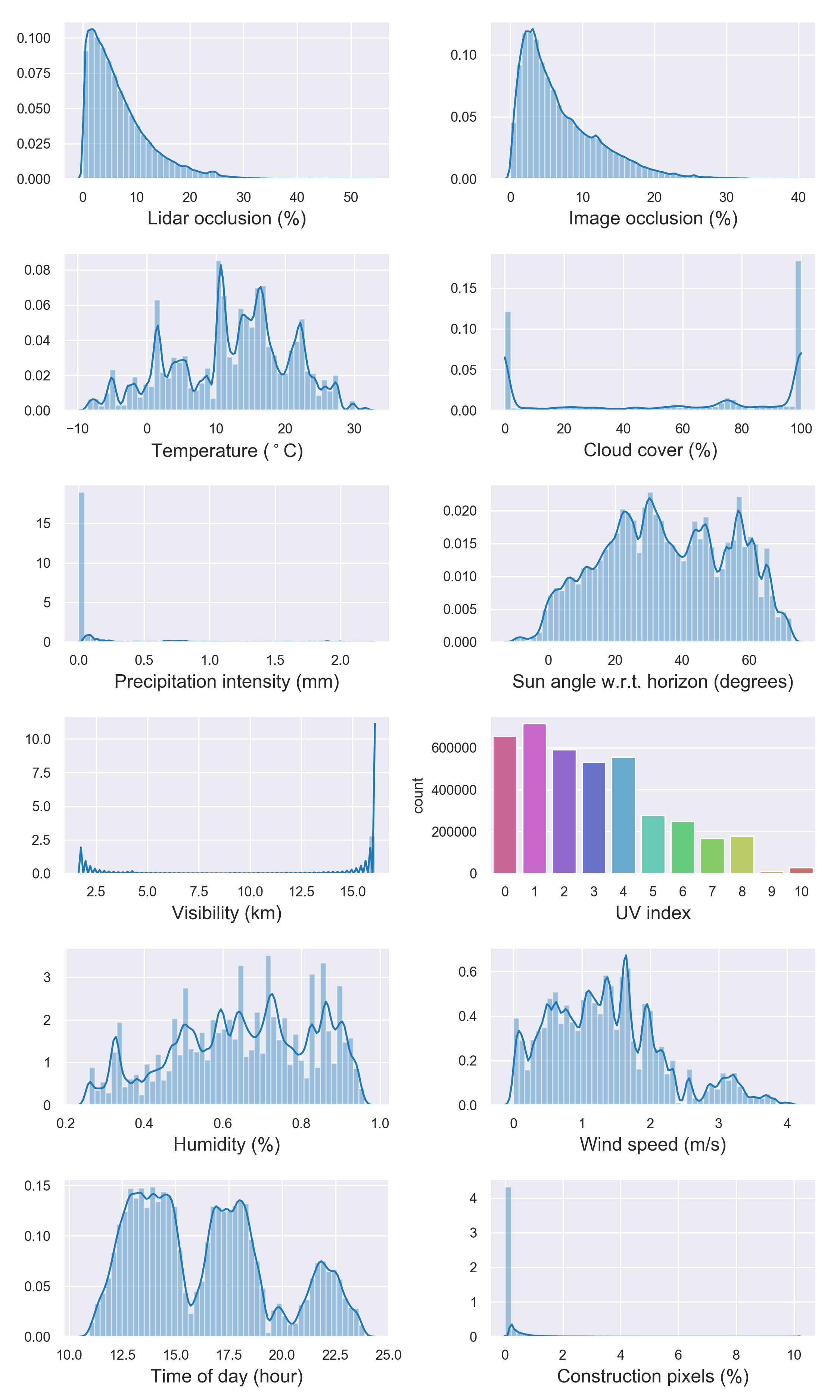}
    \caption{\textbf{Probability distribution functions of the semantic labels in Pit30M.}}
    \label{fig:labels}
\end{figure*}

As mentioned in the main paper, we provide a series of semantic labels to help researchers better understand our data.
We provide a detailed description of the data we collected in Table~\ref{tab:labels}.
We also show more plots of the distributions that we observe in our labels in Figure~\ref{fig:labels}.
For example, we observe a bimodal distribution for cloud cover and visibility,
and see that construction is present in a small fraction of the data.
We also observe Poisson-like distributions for both image and LiDAR occlusion.

\begin{table*}[h]
    \small
    \centering
    \begin{tabular}{llp{9.5cm}}
    \toprule
    Field & Units & Description \\
    \midrule
    Time of day & Datetime & Unix timestamp.\\
    Angle of the sun \wrt horizon & Degrees & Angle between the horizon and the centre of the sun, as observed from the
    location of particular sensor reading.
    This value is zero at both dawn and twilight, and is commonly used to estimate light for navigation purposes in \eg, nautical applications.\\
    \midrule
    Precipitation & Millimetres & An amount larger than zero means that either snow or rain is present.\\
    Visibility & Kilometres & Distance up to which objects can be clearly discerned. Heavy fog may result in low visibility.\\
    UV index & Integer & Integer value indicating the intensity of solar UV rays.\\
    Temperature & Degrees Celsius & Ambient temperature.\\
    Humidity & Percent & The amount of water vapour present in the air, as a percent of the maximum that the air could hold at the same temperature.\\
    Cloud cover & Percent & Percent of the sky covered by clouds.\\
    Wind speed & Metres per second & Speed of the wind.\\
    \midrule
    Image occlusion & Percent & Percent of pixels taken by dynamic objects in the scene.\\
    LiDAR occlusion & Percent & Percent of LiDAR points taken by dynamic objects in the scene.\\
    Construction & Percent & Percent of image pixels that correspond to active construction elements.\\
    \bottomrule
    \end{tabular}
    \vspace{1mm}
    \caption{{\bf Semantic labels in Pit30M.} We provide these labels to help researchers categorize and understand the performance of localization algorithms.}
    \label{tab:labels}
\end{table*}

\section{Bird's Eye View (BEV) Representation}

These are details necessary to reproduce our BEV representation, as shown in Figure~\ref{fig:lidar} (c).

\noindent {\bf Oxford Robotcar dataset:}
We preprocess the LiDAR with a resolution of 8 pixels/metre in width and height,
in an area of 40m for the x axis and 25m in the y axis centred approximately around the position of the car in the middle frame.
Along the z axis, we discretize 10m into 16 channels, resulting in a BEV image with $320 \times 200 \times 16$ voxels.
We obtain the intensity for each voxel by averaging the intensities of all the LiDAR points within the voxel.
Finally, we pass these BEV images through a standard Resnet-50 architecture~\cite{he2016deep},
except that we use max, instead of average pooling in the final layer.

\noindent {\bf Pit30M dataset:}
We use different parameters in this dataset, as the LiDAR sensors produces a 360$^\circ$ pointcloud around the car
with larger reach than the 2d sensors used in the Oxford dataset,
and the car itself casts a large shadow.
Here, we discretize an area of 200m in the x axis and 125 m in the y axis, with a resolution of 2.4 metres per pixel.
We discretize 3.2 metres in the z axis into 16 channels. We use a ResNet-50 with average pooling and follow
the training procedure as described in the main paper.

\section{Qualitative results}
\begin{figure*}[h]
  \centering
  \includegraphics[width=1\linewidth,trim=0mm 0mm 0mm 0mm,clip=true]{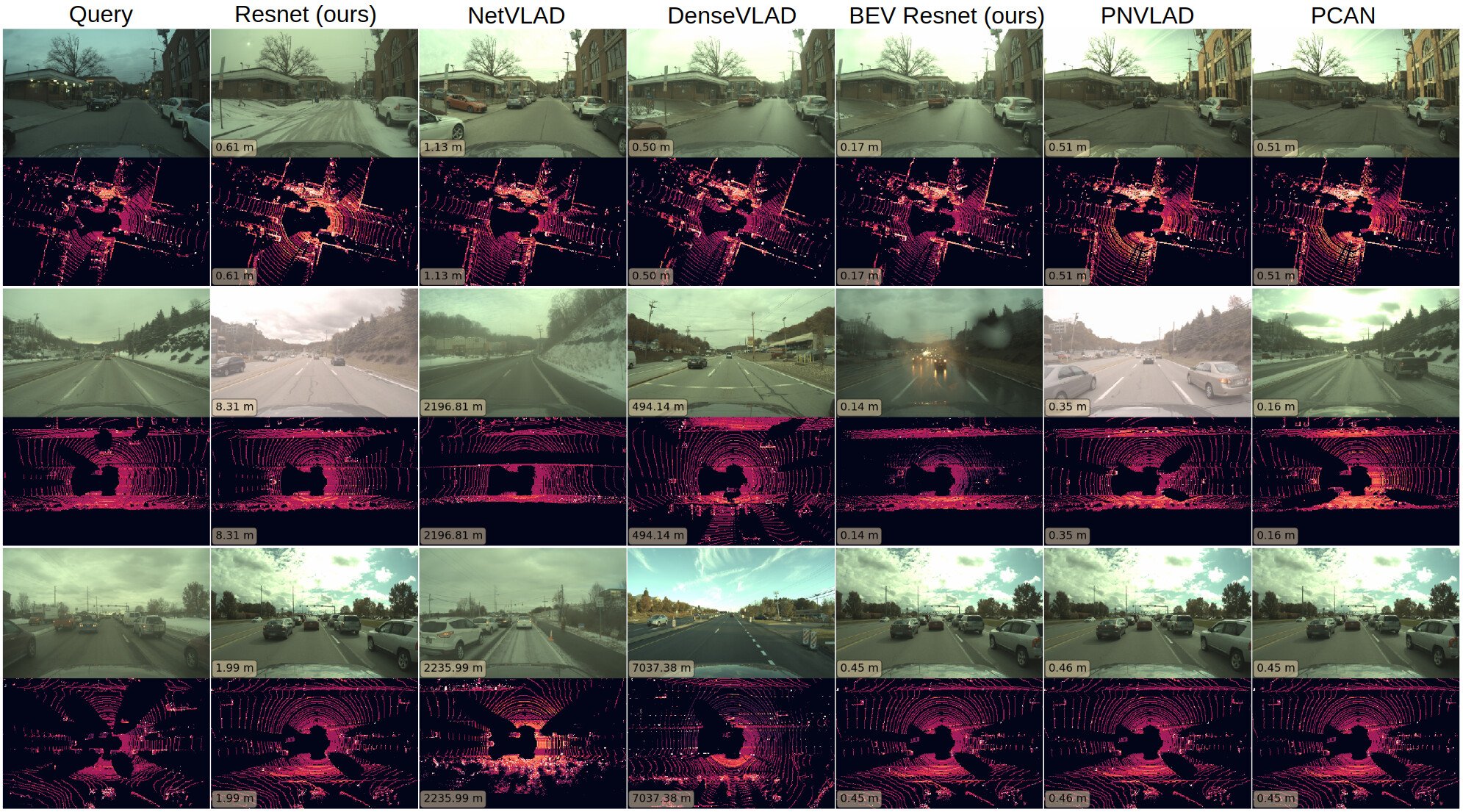}
  \caption{\textbf{Results with snow.} The second and the third example show NetVLAD and DenseVLAD struggling with
  cross-seasonal matches.}
  \label{fig:snow}
\end{figure*}

\begin{figure*}[h]
  \centering
  \includegraphics[width=1\linewidth,trim=0mm 0mm 0mm 0mm,clip=true]{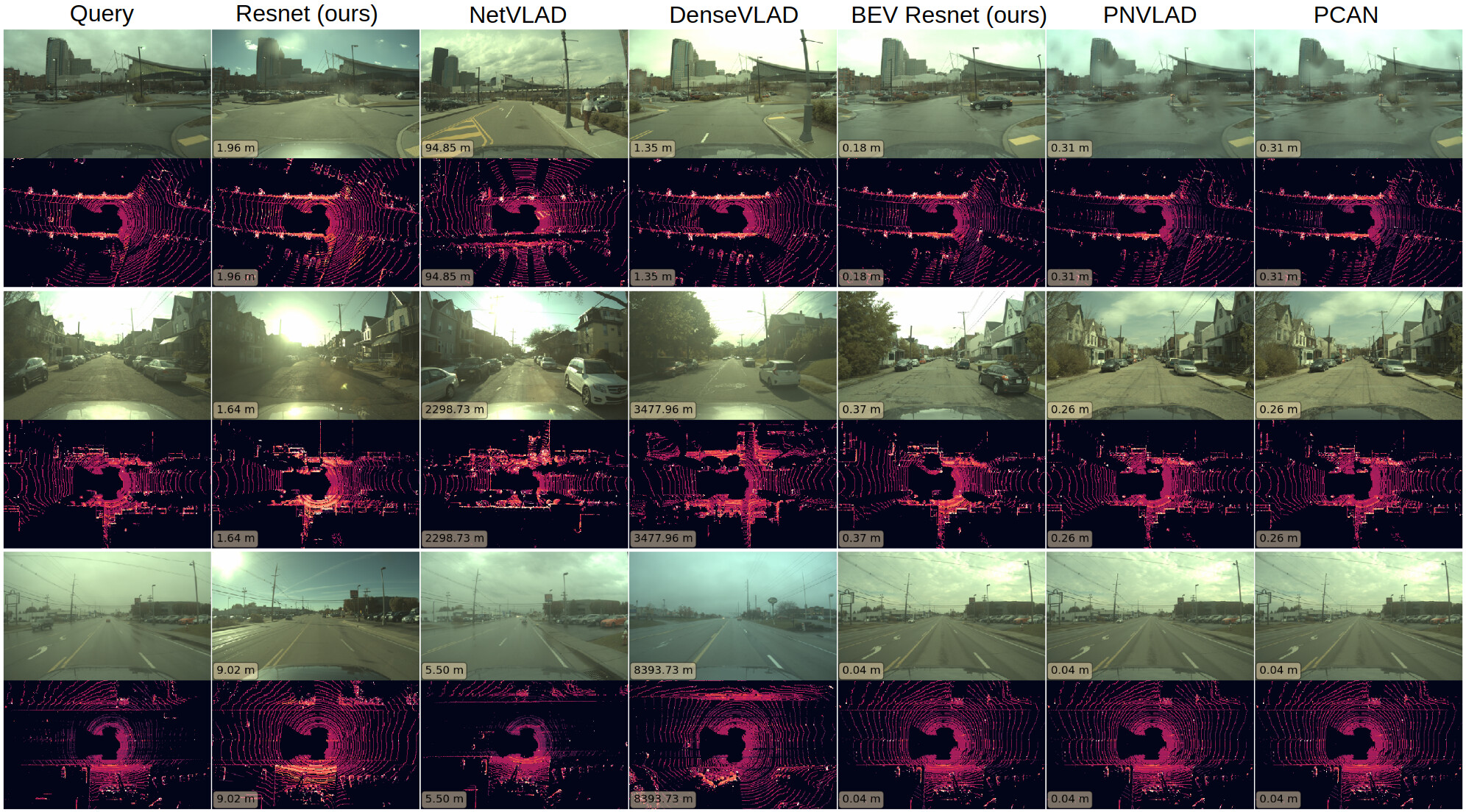}
  \caption{\textbf{Results with low sun angle.} The second example shows a successful ResNet match,
  despite the low sun clearly visible in the frame.}
  \label{fig:sun}
\end{figure*}

\begin{figure*}[h]
  \centering
  \includegraphics[width=1\linewidth,trim=0mm 0mm 0mm 0mm,clip=true]{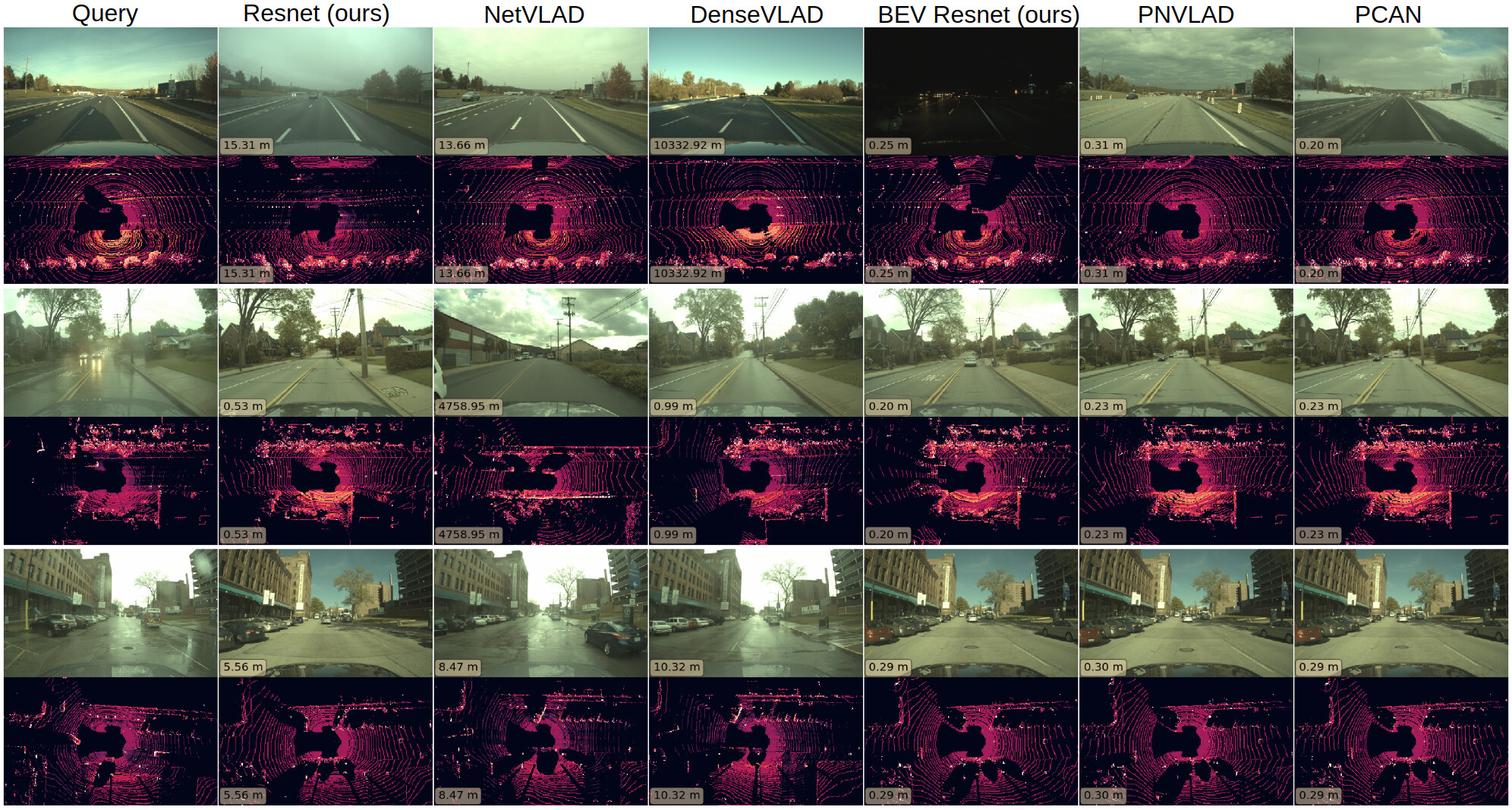}
  \caption{\textbf{Results with rain.} In the second and third examples, we see that the heavy rain in the query
  affects the matching quality in the image networks.}
  \label{fig:rain}
\end{figure*}

\begin{figure*}[h]
  \centering
  \includegraphics[width=1\linewidth,trim=0mm 0mm 0mm 0mm,clip=true]{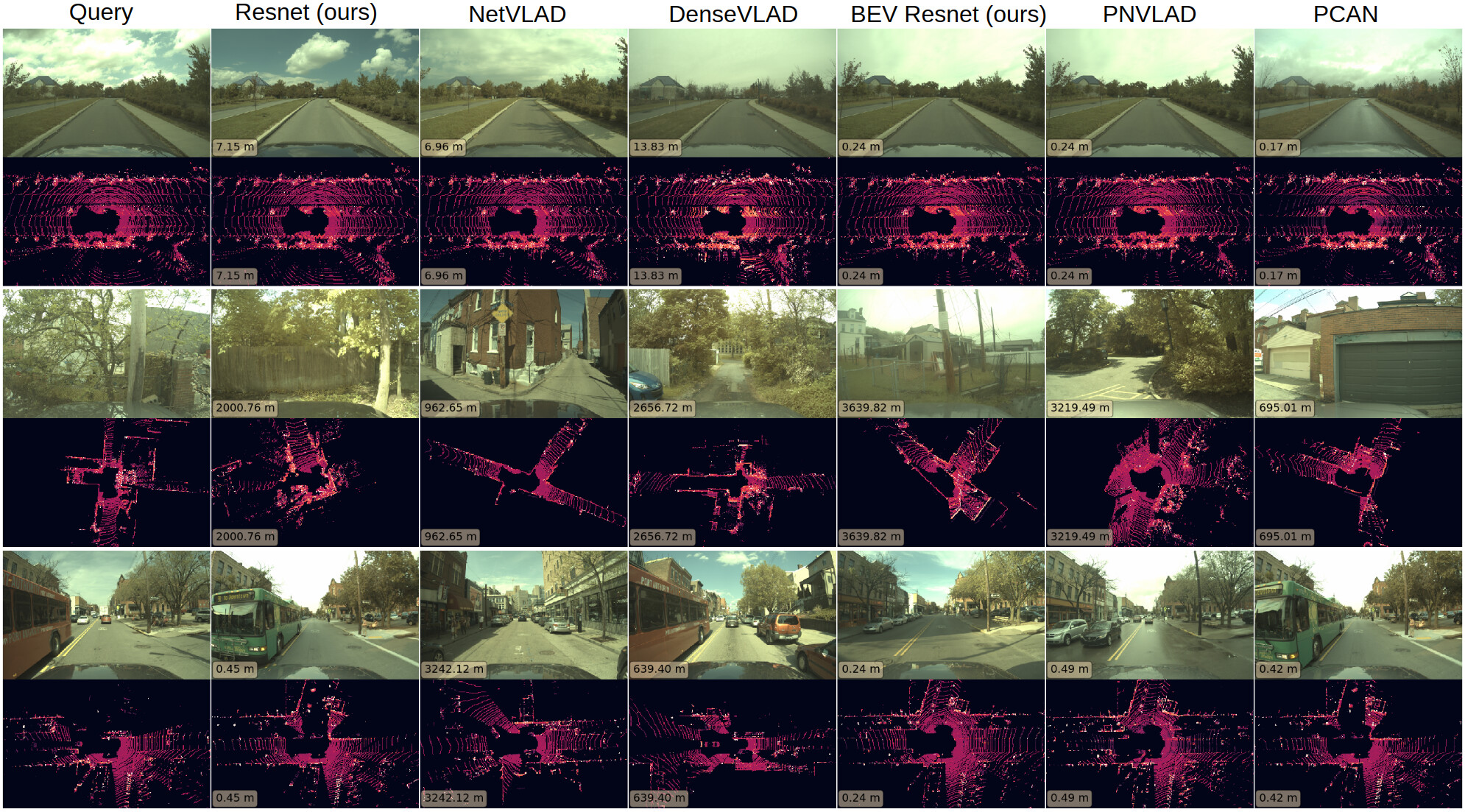}
  \caption{\textbf{Results with occlusion.} The first example demonstrates the difficulty in precise image retrieval
  with few landmarks in the image. The second example shows retrieval failures across the image and LiDAR networks,
  likely caused by the atypical location and heavy vegetation.  }
  \label{fig:occlusion}
\end{figure*}

\begin{figure*}[h]
  \centering
  \includegraphics[width=1\linewidth,trim=0mm 0mm 0mm 0mm,clip=true]{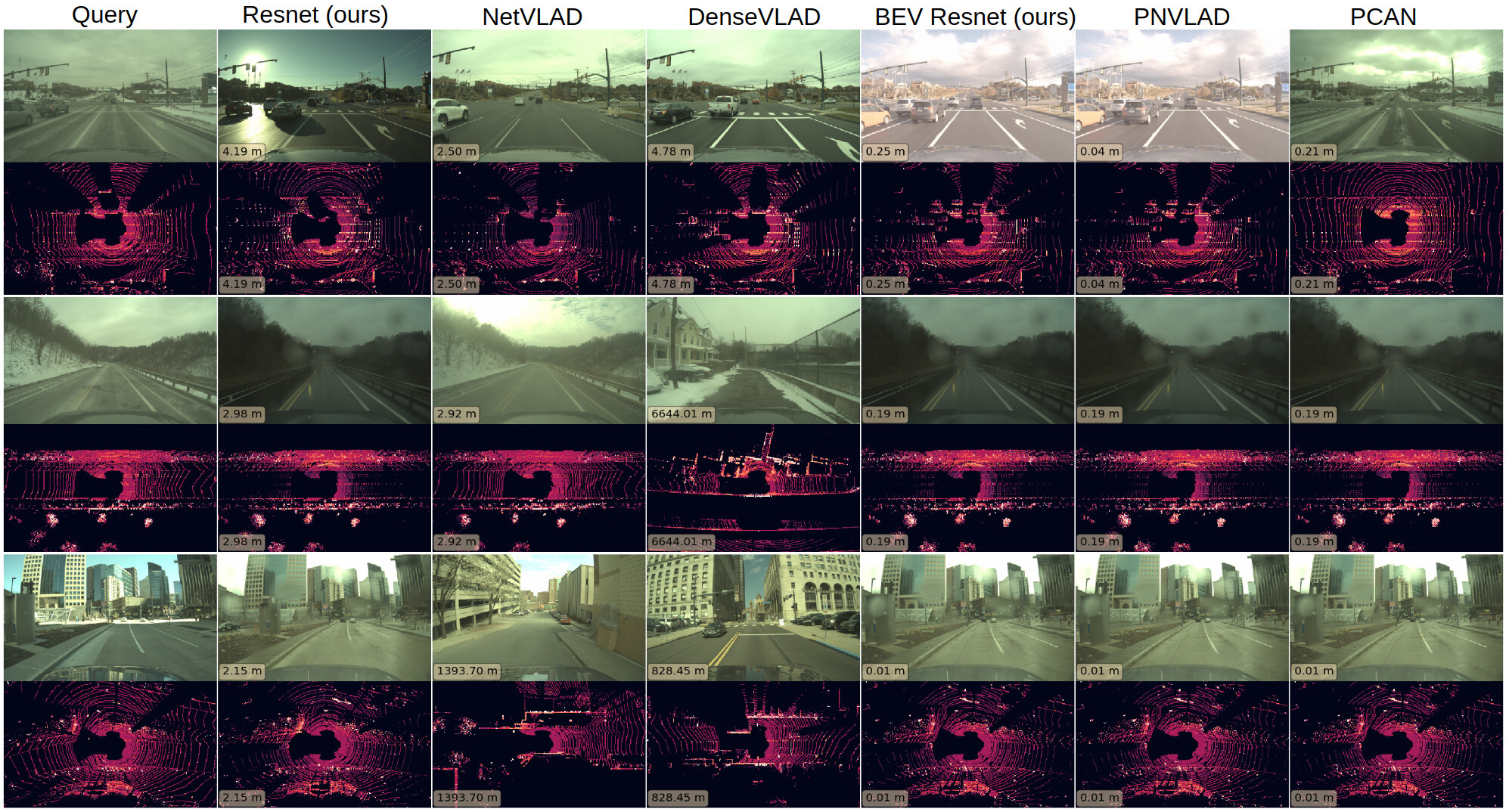}
  \caption{\textbf{Results with multiple challenging modalities.} The first example shows a low-light and snow query,
  while the last example shows both rain and sunshine, which NetVLAD and DenseVLAD have trouble handling.}
  \label{fig:combined}
\end{figure*}

\begin{figure*}[h]
  \centering
  \includegraphics[width=1\linewidth,trim=0mm 0mm 0mm 0mm,clip=true]{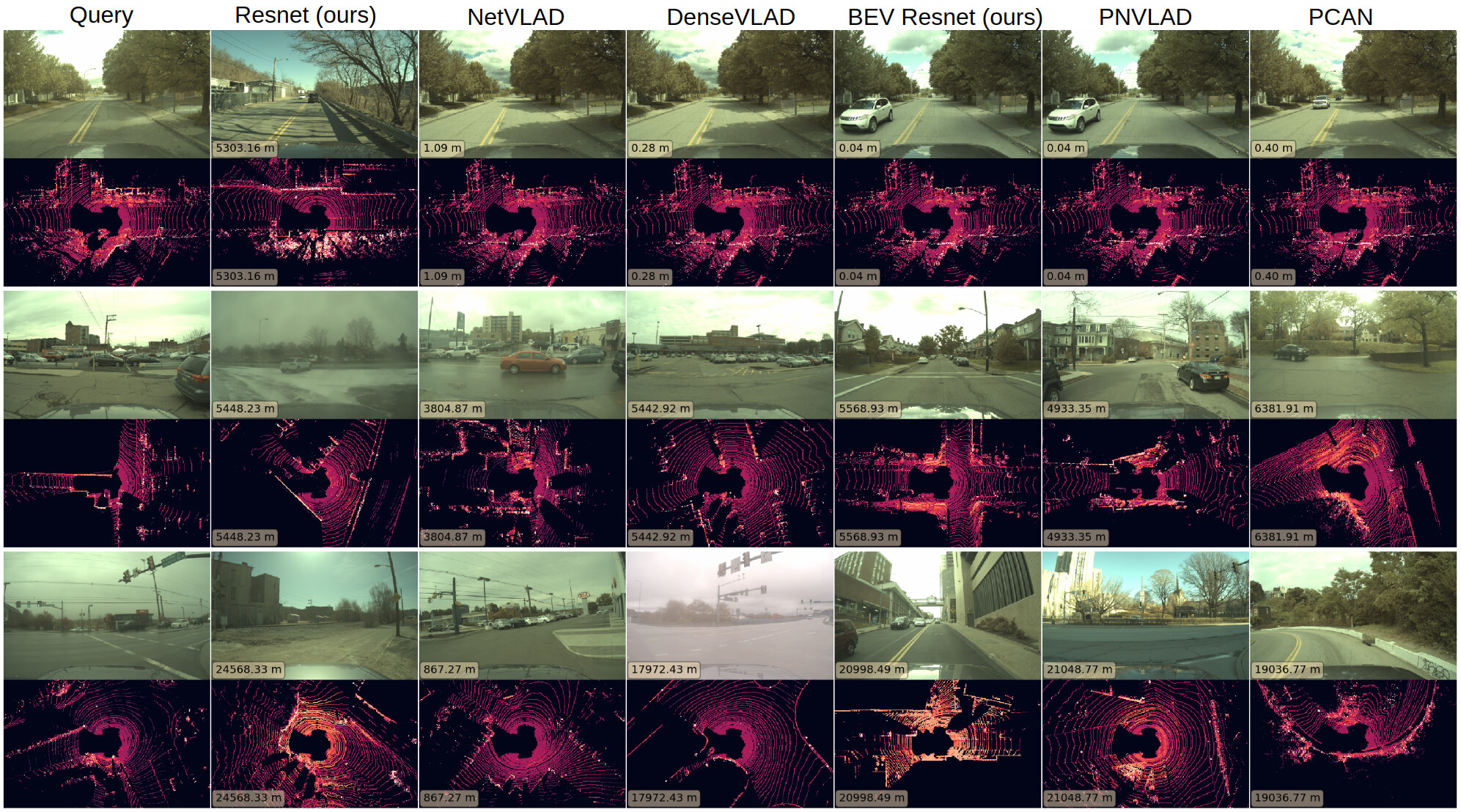}
  \caption{\textbf{Failure cases.} These second and third examples cause failures in both LiDAR and image retrieval,
  presumably due to the lack of distinctive landmarks in the sensor readings.}
  \label{fig:failure}
\end{figure*}

In Figures~\ref{fig:snow} to~\ref{fig:failure}, we show results with challenging conditions for localization, such as
images with snow, rain, low sun angle, and high occlusion. We also show some failure modes for both images and LiDAR.

\FloatBarrier
\clearpage %

\addtolength{\textheight}{-16.5cm}   %

{\small
\bibliographystyle{IEEEtran}
\bibliography{IEEEexample}
}

\end{document}